\documentclass{article} 
\usepackage{iclr2025_conference,times}

\usepackage{amsmath,amsfonts,bm}

\def\eqref#1{equation~\ref{#1}}

\def\1{\bm{1}}

\DeclareMathAlphabet{\mathsfit}{\encodingdefault}{\sfdefault}{m}{sl}
\SetMathAlphabet{\mathsfit}{bold}{\encodingdefault}{\sfdefault}{bx}{n}

\usepackage{hyperref}
\usepackage{url}
\usepackage{graphicx}

\usepackage{algorithm}%

\usepackage[frozencache,cachedir=./minted-cache]{minted}

\usepackage{listings}%
\usepackage{fvextra}    
\usepackage{lmodern}

\usepackage{booktabs} 

\title{Large Language Models and Synthetic Data for Monitoring Dataset Mentions in Research Papers}

\author{Aivin V. Solatorio \thanks{GitHub/HF: @avsolatorio, avsolatorio@gmail.com} , Rafael Macalaba \thanks{GitHub: @rafmacalaba, rafael.macalaba@yahoo.com} , \& James Liounis \thanks{ GitHub: @jamesliounis, liounisjames@gmail.com } \\
Office of the Chief Statistician \\
The World Bank \\
1818 H Street N.W., \\
Washington, 20433 \\
District of Columbia, USA \\
\texttt{\{asolatorio,rmacalaba,jliounis\}@worldbank.org} 
}

\iclrfinaltrue
\begin{document}

\maketitle

\begin{abstract}
Tracking how data is mentioned and used in research papers provides critical insights for improving data discoverability, quality, and production. However, manually identifying and classifying dataset mentions across vast academic literature is resource-intensive and not scalable. This paper presents a machine learning framework that automates dataset mention detection across research domains by leveraging large language models (LLMs), synthetic data, and a two-stage fine-tuning process. We employ zero-shot extraction from research papers, an LLM-as-a-Judge for quality assessment, and a reasoning agent for refinement to generate a weakly supervised synthetic dataset. The Phi-3.5-mini instruct model is pre-fine-tuned on this dataset, followed by fine-tuning on a manually annotated subset. At inference, a ModernBERT-based classifier efficiently filters dataset mentions, reducing computational overhead while maintaining high recall. Evaluated on a held-out manually annotated sample, our fine-tuned model outperforms NuExtract-v1.5 and GLiNER-large-v2.1 in dataset extraction accuracy. Our results highlight how LLM-generated synthetic data can effectively address training data scarcity, improving generalization in low-resource settings. This framework offers a pathway toward scalable monitoring of dataset usage, enhancing transparency, and supporting researchers, funders, and policymakers in identifying data gaps and strengthening data accessibility for informed decision-making.
\end{abstract}

\section{Introduction}

Datasets are fundamental to scientific research, underpinning empirical analysis, model development, and policy decisions \citep{mooney_anatomy_2012,stacy2024}. However, tracking how datasets are mentioned and used in academic literature remains a significant challenge \citep{potok2022show, silvello_theory_2018,stacy2024}. Unlike traditional bibliographic citations, dataset references are often embedded within text, described inconsistently, or omitted entirely, making it difficult to assess data reuse, transparency, and accessibility \citep{silvello_theory_2018,buneman_why_2020}. The lack of systematic dataset tracking limits efforts to evaluate the impact of datasets, identify underutilized resources, and address data gaps in research \citep{piwowar_data_2013,buneman_data_2021}. Without structured metadata and comprehensive monitoring, researchers, funders, and policymakers struggle to make informed decisions about data investments, availability, and governance.

Manual efforts to extract and classify dataset mentions across large volumes of scientific literature are not scalable, requiring significant time and domain expertise \citep{potok2022show,stacy2024}. Advances in artificial intelligence (AI) and natural language processing (NLP) provide a promising solution, but dataset mention detection remains underdeveloped due to the lack of annotated training data. Still, previous attempts have been made to make this type of data available \citep{Heddes2021,potok2022show,stacy2024}. Unlike traditional citation databases, which benefit from well-defined metadata, dataset references appear in unstructured formats, requiring machine learning (ML) models capable of handling diverse linguistic patterns \citep{potok2022show,hussain_novel_2023, younes_question_2023}. Addressing this challenge requires a scalable and adaptable approach that can operate effectively in low-resource settings, where labeled datasets are limited or unavailable.

This paper presents a machine learning framework that automates dataset mention detection by combining large language models (LLMs), synthetic data, and specialized classifiers. A major barrier to automating dataset tracking is the scarcity of annotated training data, as dataset mentions in research papers are highly variable and inconsistently formatted. To address this, we employ a weakly supervised learning approach that leverages LLM-generated synthetic data to bridge data gaps. Our pipeline extracts dataset mentions using zero-shot extraction \citep{kojima_large_2022}, refines outputs via an LLM-as-a-Judge \citep{gu_survey_2025} for quality assessment, and further enhances accuracy through a reasoning agent. The resulting weakly supervised dataset serves as pre-training material for the Phi-3.5-mini instruct model \citep{abdin_phi-3_2024}, which is then fine-tuned on a smaller, manually annotated subset for improved precision. At inference, a ModernBERT-based \citep{warner_smarter_2024} classifier filters dataset mentions, optimizing computational efficiency.

Our results demonstrate that LLM-generated synthetic data can effectively address data gaps, improving model robustness in low-resource scenarios and enhancing dataset mention detection. Evaluated on a held-out manually annotated sample, our fine-tuned model outperforms NuExtract-v1.5 \citep{cripwell2024} and GLiNER-large-v2.1 \citep{zaratiana_gliner_2023}, achieving state-of-the-art performance in dataset extraction. By enabling scalable monitoring of dataset usage, this framework enhances transparency, identifies data gaps, and supports efforts to improve data discoverability and accessibility. More broadly, our approach highlights the potential of synthetic data in information extraction, contributing to advancements in data use monitoring, data governance, and responsible data-sharing practices.

\begin{figure}[t]\label{fig:diagram}
\begin{center}
\includegraphics[width=\linewidth]{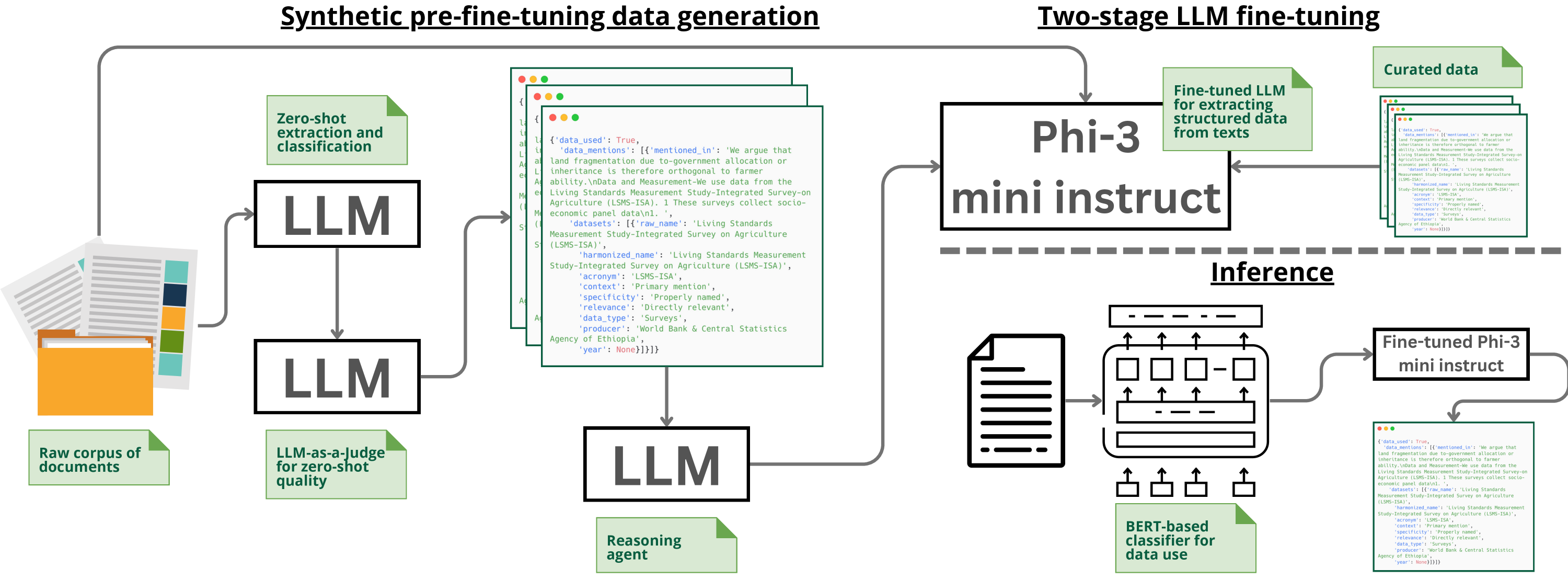}

\end{center}
\caption{Diagram for the proposed data use extraction and classification framework. The figure shows the high-level process for generating the synthetic pre-fine-tuning data, the two-stage fine-tuning of the Phi-3.5-mini instruct model, and the inference stage where ModernBERT is used to distinguish which texts likely have mentions of data and require fine-tuned LLM processing.}
\end{figure}

\section{Methods} 

This section presents our methodology for monitoring dataset mentions in research papers, with a focus on climate change literature. To address the scarcity of labeled training data for dataset extraction and classification, we leverage synthetic data generation to create a weakly supervised dataset. We then develop and fine-tune specialized models for classification and extraction, as outlined in Figure~\ref{fig:diagram}.

\subsection{Data}

Climate change research relies on diverse datasets from various sources, including weather and oceanographic data, socioeconomic indicators, satellite imagery, land-use records, greenhouse gas emissions inventories, and agricultural and biodiversity surveys \citep{meehl_wcrp_2007,camarillo-naranjo_global_2019,hasegawa_global_2022}. A corpus encompassing climate change research, therefore, provides a robust test of the method's generalizability in identifying mentions of datasets across different types and domains.

To implement and evaluate our framework, we compile a collection of climate-related research papers from two primary sources: (1) the One Earth corpus, as identified in \cite{sietsma_machine_2024}, and (2) climate-related papers from the World Bank’s Policy Research Working Papers (PRWP) series. The PRWP series encompasses a broad range of socio-economic development topics, including climate change.

To curate the PRWP subset, we use the World Bank’s Documents and Reports platform \citep{world_bank_documents_2025}, applying climate-related tags to filter relevant documents. Since full-text access is required for dataset mention extraction, we utilize the Semantic Scholar Paper Title Search API to identify open-access papers and retrieve their PDF links. This process yielded 2,123 papers with PDFs from the One Earth corpus and 582 papers from the PRWP collection. Additional information related to data acquisition and processing is described in Annex~\ref{ax:datasources}.

\subsection{Weakly Supervised Synthetic Data Generation for Pre-fine-tuning}

The lack of publicly available datasets designed to classify dataset mentions by purpose and citation quality presents a major challenge for developing reliable machine learning models for this task. While the \textit{Coleridge Initiative's Show US the Data} \citep{potok2022show} dataset provides examples of dataset mentions, it has two key limitations: \textbf{(1)} it lacks contextual information about dataset usage, making it difficult to determine how datasets are used in research, and \textbf{(2)} it focuses on a limited subset of datasets, introducing systematic bias and reducing generalizability to a wider range of research domains. This scarcity of high-quality training data poses a bottleneck in building scalable and domain-adaptive dataset extraction models.  

To address this, we leverage LLMs for extractive, classification, and pseudo-reasoning tasks to generate a weakly supervised synthetic fine-tuning dataset from our research corpus. By using LLM-generated synthetic data, we significantly reduce reliance on manually annotated datasets, making it possible to expand dataset mention detection across diverse fields where labeled training data is scarce. To ensure the models get trained with high-quality training data, we strategically sample and manually annotate portions of the weakly supervised dataset, balancing automation with human validation.  

\subsubsection{Zero-Shot Extraction and Classification}

To construct the weakly supervised pre-fine-tuning dataset, we process each page of research papers using an LLM-based extractor. This model identifies whether a dataset is mentioned and provides structured information, including:

\begin{itemize}
    \item The \textbf{dataset name} (if present).
    \item A classification of its \textbf{citation quality} (e.g., whether the dataset is explicitly named or only generically described).
    \item The \textbf{usage context}, distinguishing between datasets used for analysis, background references, or methodological descriptions.
\end{itemize}

Unlike rule-based methods, this LLM-driven approach adapts to variability in dataset citation styles, allowing it to generalize across different research papers without requiring domain-specific heuristics. However, initial extractions contain false positives, necessitating further quality control mechanisms.  

\subsubsection{LLM-as-a-Judge for Quality Assessment}

A manual review of the extracted dataset mentions revealed frequent misclassifications, where non-dataset entities (e.g., institutions, reports, or software) were incorrectly labeled as datasets. To mitigate this issue, we integrate an LLM-as-a-Judge \citep{gu_survey_2025} mechanism, where a second LLM evaluates the extracted dataset mentions for accuracy and relevance. This secondary assessment improves classification reliability by filtering out false positives before the dataset is used for training.  

While this step improves precision, LLM-based judgments are still susceptible to subtle misclassifications, particularly when dataset names resemble organization names or when datasets are ambiguously referenced. Further refinement is needed to enhance classification robustness.  

\subsubsection{Autonomous Reasoning for Filtering}

Further analysis showed that even after LLM-as-a-Judge validation, many non-dataset references persisted, including reports, conceptual frameworks, and organizations mistakenly classified as datasets. To improve classification accuracy, we introduce a \textbf{reasoning agent} that {autonomously develops and executes a structured self-evaluation strategy}. This agent systematically reassesses its conclusions, incorporating:  

\begin{itemize}
    \item A \textbf{“devil’s advocate” mechanism}, challenging its own classifications by considering alternative interpretations.
    \item A \textbf{hierarchical decision process}, where the agent re-evaluates ambiguous cases by cross-referencing multiple extraction criteria.
    \item The ability to \textbf{override previous LLM-based judgments}, provided it justifies any changes.
\end{itemize}

This dynamic self-correction process reduces reliance on implicit assumptions, enforcing stricter classification criteria and improving the reliability of dataset mention identification. The impact of this refinement was significant: out of the 37,225 mentions initially shortlisted by the LLM judge, the reasoning agent filtered out approximately 42\%, leaving 21,408 dataset mentions as likely valid.  

All LLM-based methods in this study use the OpenAI GPT-4o-mini (2024-07-18) model. The prompts used for these processes are provided in Appendices \ref{ax:zeroshotprompt}, \ref{ax:llmasajudgeprompt}, and \ref{ax:reasoningagentprompt}, corresponding to the zero-shot extraction, LLM-as-a-Judge, and reasoning agent methods, respectively.

\subsection{Fine-tuning Data}

We sampled 1,000 pages from the output of the previous method and manually annotated them using Doccano \citep{doccano} to remove any remaining false positives from the weakly supervised approach. This curated dataset serves as a foundational resource for training and evaluating downstream models specialized in extracting and classifying dataset mentions offline. We split the annotated data into three partitions for training (n=864), validation (n=40), and testing (n=20).

\subsection{Fine-tuning Models}  

\paragraph{Detecting Data Use.} To improve efficiency in dataset mention detection, we fine-tune both BERT \citep{devlin_bert_2019} and ModernBERT \citep{warner_smarter_2024} models and compare their performance. The objective is to shift the filtering stage to a lightweight encoder-based model, reducing computational overhead when processing large volumes of text. Given that dataset mentions are typically sparse within documents, our approach ensures that only passages identified as containing dataset mentions by the encoder models are passed to the LLM for further processing. This hierarchical filtering strategy optimizes resource usage while maintaining extraction accuracy. Since dataset mentions are first filtered by an encoder model before being passed to the LLM for extraction, there is a possibility that some datasets may not be identified if the encoder misclassifies the passage containing the mention. To mitigate this, we need to optimize and select a model that possesses the highest recall.

\paragraph{Extracting and Classifying Dataset Mentions.}  
We fine-tune the Phi-3.5-mini instruct model \citep{abdin_phi-3_2024} using 16-rank LoRA \citep{hu_lora_2021} to extract structured information about dataset mentions in text. Our training follows a two-stage fine-tuning approach:  

\begin{itemize}
    \item \textit{Pre-fine-tuning on synthetic data} – We first pre-fine-tune the model using our large weakly supervised synthetic dataset for 10 epochs with an effective batch size = 16, learning rate = 2e-4, warmup ratio of 1\%, and linear scheduler with decay of 0.01, saving a checkpoint every 100 steps and tracking the best-performing model based on validation loss.
    \item \textit{Fine-tuning on high-quality annotated data} – After pre-fine-tuning, we load the best-performing checkpoint and further fine-tune the model on a smaller, manually annotated dataset with the same configurations as the pre-fine-tuning except for it training over 20 epochs, an effective batch size = 2, and learning rate = 3e-5, with checkpoints saved every 50 steps (and created at the end of each epoch) and the model corresponding to the best-performing evaluation loss is saved and loaded after training.
\end{itemize}  

This two-stage fine-tuning strategy enables the model to establish a broad representation of dataset mentions using the diverse, albeit imperfect, synthetic corpus before refining its understanding on the manually curated dataset. By progressively adapting to more precise annotations, we expect the model to enhance its ability to distinguish between dataset references and similar entities, improving classification robustness.  We perform an ablation study to assess the impact of each fine-tuning stage.

To evaluate our approach, we compare performance against NuExtract-v1.5 \citep{cripwell2024} and GLiNER-large-v2.1 \citep{zaratiana_gliner_2023}, two state-of-the-art models for named entity recognition and zero-shot structured data extraction, serving as baselines.

\section{Results}  

\vspace{-0.15cm}

\begin{table}[t]
\vspace{-0.4cm}
\caption{Performance of Classification and Extraction Models for Data Use}
\vspace{0.25cm}
\label{datause-models}
\renewcommand{\arraystretch}{1} 
\small 
\centering
\begin{tabular}{llll}
\toprule
\multicolumn{4}{c}{\textbf{Data Use Classification Models}} \\ 
\midrule
\textbf{Model}  &\textbf{Precision}  &\textbf{Recall} &\textbf{F1-score} \\
\cmidrule(lr){1-4}
BERT-uncased         &\textbf{100.0} &50.0 &67.0 \\
ModernBERT-base     &\textbf{100.0} &\textbf{100.0} &\textbf{100.0} \\
\midrule
\multicolumn{4}{c}{\textbf{Data Use Extraction Models}} \\ 
\midrule
\textbf{Model [Training data]}  &\textbf{Precision}  &\textbf{Recall} &\textbf{F$\beta$-score} \\
\cmidrule(lr){1-4}
Phi-3-mini [Synthetic and curated]  &\textbf{69.45} &\textbf{80.65} &\textbf{71.43} \\
Phi-3-mini [Synthetic only]  &60.00 &70.00 &61.76 \\
Phi-3-mini [Curated only]  &55.68 &65.52 &57.58 \\
GLiNER-large-v2.1       &62.50 &71.43 &64.10 \\
NuExtract-v1.5       &20.97 &46.43 &23.55 \\
\bottomrule
\end{tabular}
\end{table}

Our evaluation assesses both classification and extraction models for detecting dataset mentions in research papers. Table~\ref{datause-models} summarizes the performance of the models across different training configurations and baselines.

\vspace{-0.1cm}
\subsection{Impact of Fine-Tuning on Extraction Performance}  
\vspace{-0.1cm}

To measure the extraction accuracy, we use the Jaccard Similarity-based (Equation~\ref{eq:jaccard}) F$\beta$ Score, a metric introduced in the Coleridge Initiative's data extraction Kaggle competition \citep{gupta_evaluation_2021}. This method evaluates the overlap between predicted and ground-truth dataset mentions, allowing for partial matches rather than requiring exact string matches—an important consideration given the variability in how datasets are referenced in text and how the models select which snippets constitute dataset names.

\begin{equation}
J(S_1, S_2) = \frac{|W_1 \cap W_2|}{|W_1| + |W_2| - |W_1 \cap W_2|}
\label{eq:jaccard}
\end{equation}

In Equation~\eqref{eq:jaccard}, \( J(S_1, S_2) \) represents the Jaccard similarity between two strings \( S_1 \) and \( S_2 \). The sets \( W_1 \) and \( W_2 \) contain the unique words obtained from tokenizing \( S_1 \) and \( S_2 \), respectively. The term \( |W_1 \cap W_2| \) denotes the number of words common to both sets, while \( |W_1| \) and \( |W_2| \) represent the total number of unique words in each set. The denominator, \( |W_1| + |W_2| - |W_1 \cap W_2| \), ensures that words shared between both sets are not double-counted, yielding a similarity score between 0 and 1. In this context, a Jaccard score greater than 0.5 is considered a match. Based on this classification, the precision, recall, and F\(\beta\)-score are computed to evaluate performance.



Our results show that fine-tuning the Phi-3-mini instruct model significantly improves extraction accuracy, outperforming NuExtract-v1.5 and GLiNER-large-v2.1. Among the different fine-tuning strategies:  

\vspace{-0.1cm}

\begin{itemize}
    \item The Phi-3-mini model trained on both synthetic and curated data achieves the highest F$\beta$ score (71.43), demonstrating the effectiveness of a two-stage fine-tuning approach: pre-fine-tuning with a larger synthetic dataset followed by refinement on a smaller, manually curated sample. This combination improves generalization while enhancing precision.
    \item The synthetic-only model outperforms the curated-only model, achieving higher recall (70.00 vs. 65.52), suggesting that synthetic data, despite lacking human-verified precision, contributes to broader coverage and generalization. However, the curated-only model offers better precision, reinforcing the importance of human-verified refinements.
    \item The baseline NuExtract-v1.5 model performs significantly worse, while GLiNER-large-v2.1 achieves comparable results to the LLM pre-fine-tuned exclusively on synthetic data. However, both underperform relative to the two-stage fine-tuned LLM, underscoring the advantages of domain-adapted fine-tuning.
\end{itemize}  

These findings highlight the benefits of leveraging synthetic data for broad generalization while using curated data to refine accuracy. The 9.67-point improvement in F$\beta$ score with curated fine-tuning demonstrates its critical role in enhancing dataset mention extraction. This also suggests that further optimizing the pre-fine-tuned LLM with topic-specific annotated data---such as curated dataset mentions in disaster management, refugee and forced displacement, or labor markets---could enhance its adaptability to specialized domains.

Our ablation study further confirms that pre-fine-tuning is crucial when only a small volume of annotated data is available. Conditioning the model with synthetic data mitigates overfitting to limited human annotations and enhances recall for unseen dataset mentions, improving adaptability across diverse research contexts.

\vspace{-0.1cm}
\subsection{Classification Model Performance}  

In addition to dataset extraction, we evaluate classification models that determine whether a text passage is likely to contain a dataset mention. This step serves as a filtering mechanism before passing text to the extraction model, improving computational efficiency.

\vspace{-0.1cm}

\begin{itemize}
    \item ModernBERT-base achieves perfect precision and recall (100.0 F1-score), making it the preferred choice for classifying dataset mentions.  
    \item BERT-uncased struggles with recall (50.0), leading to an F1-score of 67.0. This suggests that while BERT can correctly classify some dataset mentions, it frequently fails to detect them, potentially impacting recall in the extraction pipeline.  
    \item One likely reason for ModernBERT's superior performance is its ability to process 2048-token contexts, which we were able to set, compared to BERT's 512-token limitation. The larger context size allows ModernBERT to better capture dataset mentions, especially when they appear in longer textual discussions.  
\end{itemize}  

\vspace{-0.1cm}

\subsection{Discussion}  

Our key finding is that relying solely on small, manually curated datasets is suboptimal. However, these curated datasets become significantly more effective when used to fine-tune a synthetic-data-preconditioned LLM for the task. By adopting a two-stage fine-tuning strategy—first with synthetic data, then with curated data—our approach provides a scalable solution for tracking dataset usage in research papers.  

These results further highlight the benefits of pre-fine-tuning on synthetic data for improving dataset mention detection while maintaining high recall. Additionally, the performance of ModernBERT as a filtering model suggests that context length plays a critical role in classification accuracy, with larger context windows enhancing the model’s ability to capture dataset references more effectively.

In Annex~\ref{ax:empirical-predicted-table}, we provide examples of extracted dataset mentions alongside empirical dataset references, demonstrating the model's effectiveness in identifying relevant datasets.

\vspace{-0.1cm}
\section{Conclusion} 
\vspace{-0.1cm}

This paper introduced a machine learning framework for automating dataset mention detection in research papers, combining synthetic data generation with a two-stage fine-tuning approach. Our results demonstrate that pre-fine-tuning on a weakly supervised dataset before manual fine-tuning significantly improves extraction accuracy while maintaining high recall. The fine-tuned Phi-3-mini instruct model outperforms state-of-the-art baselines, highlighting the value of synthetic data in addressing training data scarcity and improving model generalization to unseen dataset mentions. Additionally, our classification experiments show that ModernBERT’s larger context window enhances filtering efficiency, reducing computational overhead while ensuring high recall.  

Beyond improving dataset discoverability, this approach contributes to scalable research monitoring and metadata generation, supporting open science initiatives and data-driven decision-making. Future work will include building larger manually annotated data and exploring the hypothesis of potential diminishing returns when pre-fine-tuning with synthetic data is employed. We will also explore improvements to generating the synthetic data as well as adaptive fine-tuning strategies to further refine extraction accuracy and expand the dataset to cover a broader range of research domains, improving the generalizability of dataset mention detection.

\subsubsection*{Acknowledgments}
This work is supported by the ``KCP IV - Exploring Data Use in the Development Economics Literature using Large Language Models (AI and LLMs)" project funded by the Knowledge for Change Program (KCP) of the World Bank - RA-P503405-RESE-TF0C3444.

\subsubsection*{Disclaimer and Disclosure of AI Use}

The findings, interpretations, and conclusions expressed in this paper are entirely those of the authors. They do not necessarily represent the views of the International Bank for Reconstruction and Development/World Bank and its affiliated organizations, or those of the Executive Directors of the World Bank or the governments they represent.

This work used AI tools at various stages, including generating synthetic data and reasoning using the gpt-4o-mini model (API) and open source AI models. In addition, Microsoft Co-Pilot and ChatGPT were employed to enhance the manuscript's readability.

\bibliography{iclr2025_conference}
\bibliographystyle{iclr2025_conference}

\newpage
\appendix
\section{Appendix}

\raggedbottom

\subsection{Data Sources}\label{ax:datasources}

The datasets/corpus used in this study are derived from:

\begin{itemize}
    \item \textbf{One Earth corpus source:} The dataset where the One Earth corpus was derived from was obtained from \href{https://zenodo.org/records/7893023}{Zenodo (\url{https://zenodo.org/records/7893023})}. The dataset was introduced in \citep{sietsma_machine_2024}, which provides context and methodological details regarding its creation.
    \item \textbf{PRWP corpus source:} To identify Policy Research Working Papers (PRWPs) relevant to climate change, a structured filtering approach was applied using the \textit{World Bank Documents and Reports} portal (\href{https://documents.worldbank.org/en/publication/documents-reports/documentlist?qterm=climate\%20change&keyword_select=allwords&subtopic_key=1070637^1070634^672763^1070635^1070636^1070646^1070641^1070642^1070643^1070654&docty_key=620265&lang_key=120701&srt=docdt&order=desc}{source}).
    \begin{itemize}
        \item The query parameters in the source URL indicate the applied filters, ensuring that only documents meeting specific criteria were selected. The filtering criteria included:
        \begin{itemize}
            \item \textbf{Document Type:} Policy Research Working Papers (PRWP)
            \item \textbf{Query:} climate change
            \item \textbf{Language:} English
            \item \textbf{Selected Topics:}
            \begin{itemize}
                \item Climate Change and Agriculture
                \item Adaptation to Climate Change
                \item Climate Change and Environment
                \item Climate Change Impacts
                \item Climate Change Mitigation and Greenhouse Gases
                \item Climate Change and Health
                \item Climate Change Economics
                \item Investment and Investment Climate
                \item Climate Change Policy and Regulation
                \item Climate and Meteorology
                \item Science of Climate Change
                \item Social Aspects of Climate Change
            \end{itemize}
        \end{itemize}
    \end{itemize}

\end{itemize}

The process for building the corpus involved ensuring a PDF is available for the paper. This requires the following approach: metadata retrieval, validating for open access, and downloading of the PDFs.

We retrieve the metadata via Semantic Scholar:

\begin{itemize}
    \item The selected paper titles were queried in \textit{Semantic Scholar} using the Paper Title Search API (\url{https://api.semanticscholar.org/graph/v1/paper/search/match}). This allowed the retrieval of structured metadata, including authorship details, publication year, abstracts, and citation counts, and, importantly, a flag indicating if the resource is open access.
    \item If available, PDFs of the identified papers were downloaded for further analysis.
\end{itemize}








\subsection{Templates and classifications}

Below is the JSON template used for extracting data mentions using the NuExtract-v1.5 model.

\begin{listing}[H]
    \begin{minted}[breaklines, breaksymbolleft={}]{text}

    NU_TEMPLATE = {
      "data_mentions": [
        {
          "mentioned_in": "",
          "datasets": [
            {
              "raw_name": "",
              "acronym": ""
            }
          ]
        }
      ]
    }
    \end{minted}
    \caption{JSON template for the NuExtract-v1.5 model.}
    \label{lst:nuextract-template}
\end{listing}

\subsection{Listing of empirical and extracted data mentions}

\begin{table}[H]
\centering
\caption{Empirical and Predicted Datasets}
\label{ax:empirical-predicted-table}
\resizebox{0.98\textwidth}{!}{
\begin{tabular}{p{7cm} p{7cm}}
\toprule
\textbf{Empirical} & \textbf{Predicted} \\ 
\midrule 

Africa Rainfall and Temperature Evaluation System (ARTES) \newline 
Soil data \newline 
Hydrology data from the University of Colorado & 
Africa Rainfall and Temperature Evaluation System (ARTES) \newline 
Soil data from FAO \newline 
Data concerning hydrology from the University of Colorado \\ 
\midrule 

India’s quinquennial labor force survey \newline 
30-year agricultural wage series for Indian districts \newline 
Wholesale crop price data & 
Domestic crop price data \newline 
Crop price data \\ 
\midrule 

Balanced-panel of 2,382 households & 
Baseline survey \\ 
\midrule 

Enquête Agricole de Conjoncture Intégrée aux Conditions de Vie des Ménages (EAC-I) \newline 
Fourth-General Census of Population and Housing (2009) \newline 
Meteorological data & 
Mali’s Enquête Agricole de Conjoncture Intégrée aux Conditions de Vie des Ménages (EAC-I) \newline 
Fourth General Census of Population and Housing (2009) \\ 
\midrule 

Household survey data \newline 
Republic of Uganda 2005 & 
- \\ 
\midrule 

Shock modules & 
Shock modules \\ 
\midrule 

2005 SAM for Ghana & 
2005 SAM for Ghana \\ 
\midrule 

Agroalimentary and Fisheries Information Service (SIAP) \newline 
Coupled Model Intercomparison Project Phase 3 (CMIP3) \newline 
Income and Expenditure Household Survey (ENIGH) \newline 
Count of Population and Housing 2005 \newline 
2007 Agricultural Census & 
Agroalimentary and Fisheries Information Service (SIAP) \newline 
National Weather Service (SMN) \newline 
National Water Commission (CONAGUA) \newline 
Income and Expenditure Household Survey (ENIGH) \newline 
Count of Population and Housing 2005 \newline 
Summary Statistics of the 2007 Agricultural Census (INEGI) \\ 
\midrule 

Africa Rainfall and Temperature Evaluation System (ARTES) \newline 
Soil data \newline 
Hydrology data from the University of Colorado & 
Africa Rainfall and Temperature Evaluation System (ARTES) \newline 
Soil data \newline 
Hydrology data \\ 
\midrule 

Climate data from the 18 meteorological stations of highest quality in Bolivia from May 1948 to May 2008 & 
Climate data from the 18 meteorological stations of highest quality in Bolivia \\ 
\midrule 

- & - \\ 
\midrule 

- & - \\ 
\midrule 

- & - \\ 
\midrule 

- & - \\ 
\midrule 

- & - \\ 
\midrule 

- & - \\ 
\midrule 

- & - \\ 
\midrule 

- & - \\ 
\midrule 

- & Toxic Release Inventory (TRI) \\ 
\bottomrule
\end{tabular}
}
\end{table}

\subsection{Prompts}

\subsubsection{Zero-shot extraction and classification prompt}\label{ax:zeroshotprompt}

\begin{listing}[H]
    \begin{minted}[breaklines, breaksymbolleft={}]{text}
    You are an expert in extracting and categorizing dataset mentions from research papers and policy documents. Your task is to **identify and extract all valid dataset mentions**, ensuring they are correctly classified based on naming specificity, context, and relevance.
    
    ### **What Qualifies as a Dataset?**
    A dataset is a structured collection of data used for empirical research, analysis, or policy-making. Examples include:
    - **Surveys & Census Data** (e.g., LSMS, DHS, national census records)
    - **Indicators & Indexes** (e.g., HDI, GFSI, WDI, ND-GAIN, EPI)
    - **Geospatial & Environmental Data** (e.g., OpenStreetMap, Sentinel-2 imagery)
    - **Economic & Trade Data** (e.g., UN Comtrade, Balance of Payments Statistics)
    - **Health & Public Safety Data** (e.g., epidemiological surveillance, crime reports)
    - **Time-Series & Energy Data** (e.g., climate projections, electricity demand records)
    - **Transport & Mobility Data** (e.g., road accident statistics, smart city traffic flow)
    - **Other emerging dataset types** as identified in the text.
    
    **Important:**  
    If the dataset does not fit into the examples above, infer the **most appropriate category** from the context and **create a new `"data_type"` if necessary**.
    
    ### **What Should NOT Be Extracted?**
    Do **not** extract mentions that do not clearly refer to a dataset, including, but not limited to:
    1. **Organizations & Institutions** (e.g., WHO, IMF, UNDP, "World Bank data" unless it explicitly refers to a dataset)
    2. **Reports & Policy Documents** (e.g., "Fiscal Monitor by the IMF", "IEA Energy Report"; only extract if the dataset itself is referenced)
    3. **Generic Mentions of Data** (e.g., "various sources", "survey results from multiple institutions")
    4. **Economic Models & Policy Frameworks** (e.g., "GDP growth projections", "macroeconomic forecasts")
    5. **Legislation & Agreements** (e.g., "Paris Agreement", "General Data Protection Regulation")

    \end{minted}
    \caption{System prompt used to extract the initial structured data containing likely data mentions from a given text [1/3].}
    \label{lst:system-prompt-extraction-01}
\end{listing}

\begin{listing}[H]
    \begin{minted}[breaklines, breaksymbolleft={}]{text}
    ### **Rules for Extraction**
    1. **Extract All Structured Data Mentions**
       - If the dataset is explicitly named (e.g., "Global Fishing Watch"), label it as `"properly_named"`.
       - If the dataset is described but not explicitly named (e.g., "electricity usage data from Albania"), label it as `"descriptive_but_unnamed"`.
       - If the dataset mention is too generic (e.g., "electricity usage data"), label it as `"vague_generic"`.
    
    2. **Ensure `"data_type"` Is Always Assigned**
       - **Use an existing category if applicable.**
       - **If no suitable category exists, create a new `"data_type"` based on context.**
    
    3. **Classify `"context"` Correctly**
       - `"primary"`: The dataset is used for direct analysis in the document.
       - `"supporting"`: The dataset is referenced to validate or compare findings.
       - `"background"`: The dataset is mentioned as general context or prior research.
    
       **Examples:**
       - `"The LSMS-ISA data is analyzed to assess the impact of agricultural practices on productivity."` → `"primary"`
       - `"Our results align with previous studies that used LSMS-ISA."` → `"supporting"`
       - `"LSMS-ISA is widely recognized as a reliable data source for agricultural research."` → `"background"`
    
    4. **Capture Full Sentence Context**
       - The `"mentioned_in"` field must always include the **full sentence** where the dataset is referenced.
       - If a dataset is mistakenly extracted from an unrelated sentence, correct it.
    \end{minted}
    \caption{[Continued] System prompt used to extract the initial structured data containing likely data mentions from a given text [2/3].}
    \label{lst:system-prompt-extraction-02}
\end{listing}

\begin{listing}[H]
    \begin{minted}[breaklines, breaksymbolleft={}]{text}
    ### **Extraction Schema**
    Each extracted dataset should have the following fields:
    - `raw_name`: Exact dataset name from the text (**no paraphrasing**).
    - `harmonized_name`: If properly named, use directly; if referenced in multiple ways, standardize using the most precise form in the text, otherwise, set this to None.
    - `acronym`: Extract if explicitly mentioned.
    - `mentioned_in`: **Full sentence** where the dataset appears (**no paraphrasing**).
    - `context`: **primary / supporting / background**
    - `specificity`: **properly_named / descriptive_but_unnamed / vague_generic**
    - `relevance`: **directly_relevant / indirectly_relevant / not_relevant**
    - `producer`: **Extract only if explicitly mentioned; otherwise, set to `None`.**
    - `data_type`: **Assign based on existing categories, but create new ones if necessary.**
    
    ### **Handling New or Unlisted Data Types**
    - If a dataset does not fit into existing categories, **infer an appropriate name** for its `"data_type"` based on context.
    - Use a **general but informative label** for new data types (e.g., `"Climate Risk Data"`, `"Social Media Analytics"`).
    
    ### **Important: Do NOT Skip Unnamed Datasets**
    If a dataset is described but lacks a proper name, extract it under `"descriptive_but_unnamed"` or `"vague_generic"`, which ever is appropriate.
    If `"producer"` is not mentioned, set it to `None` rather than inferring.

    \end{minted}
    \caption{[Continued] System prompt used to extract the initial structured data containing likely data mentions from a given text [3/3].}
    \label{lst:system-prompt-extraction-03}
\end{listing}

\subsubsection{LLM-as-a-Judge prompt}\label{ax:llmasajudgeprompt}

\begin{listing}[H]
    \begin{minted}[breaklines, breaksymbolleft={}]{text}
    You are an expert in dataset validation. Your task is to assess whether each dataset mention is **valid, invalid, or requires clarification**, ensuring correctness and consistency based on the dataset's **empirical context**.
    
    ---
    
    ### **Dataset Validation Criteria**
    A dataset is **valid** if:
    1. **It is structured**—collected systematically for research, policy, or administrative purposes.
    2. **It is reproducible**—meaning it consists of collected records rather than being derived purely from computations or models.
    
    **Always Valid Datasets:**
    - Government statistical and geospatial datasets (e.g., census, official land records).  
    - Official surveys, administrative records, economic transaction data, and scientific research datasets.  
    
    **Invalid Datasets:**
    Set as invalid all `"raw_name"` that belong under the following classes.
    - Derived indicators or computational constructs (e.g., "wealth score", "mine dummy", "district total production").  
    - Standalone statistical metrics without a clear underlying dataset (e.g., "average income growth rate" without source data).  
    - General organizations, reports, or methodologies (e.g., "World Bank", "UNDP Report", "machine learning model").  
    
    **Uncertain Cases:**
    - If a dataset is **vaguely named but potentially valid**, set it as valid but return: `"Potentially valid—needs dataset name confirmation."`  
    - If a dataset reference is **too generic** (e.g., `"time-varying data on production"`), set it as valid but return: `"Needs clarification—dataset name is too generic."`  
    \end{minted}
    \caption{System prompt used to characterize the LLM-as-a-Judge agent to assess the quality of the first stage of structured data generation [1/2].}
    \label{lst:system-prompt-llm-as-a-judge-01}
\end{listing}

\begin{listing}[H]
    \begin{minted}[breaklines, breaksymbolleft={}]{text}
    ---
    
    ### **Key Validation Rules**
    1. **Consistency Check:**  
       - If a `"raw_name"` has been marked **valid earlier**, it **must remain valid** unless its meaning significantly differs in a new context.
    
    2. **Context-Aware Inference:**  
       - If certain details are missing such as the **Year**, **Producer**, or **Data Type**, try to extract them from the `mentioned_in` field if available and correctly relate to the data.
    
    3. **Data Type Classification (Flexible & Adaptive):**  
       - Infer the most appropriate `"data_type"` dynamically from context.  
       - Possible types: **Surveys, geospatial data, administrative records, financial reports, research datasets, climate observations, etc.**  
       - If **no predefined category fits**, create a **new `"data_type"` that best describes the dataset.**  
    
    4. **Producer Identification:**  
       - If the **producer (organization/institution) is explicitly mentioned**, extract it.  
       - If not mentioned, **do not infer—set `"producer": None"` instead.**  
    
    ---
    
    ### **JudgeResponseFormat Schema**
    Each dataset assessment must conform strictly to the JudgeResponseFormat schema."

    \end{minted}
    \caption{[Continued] System prompt used to characterize the LLM-as-a-Judge agent to assess the quality of the first stage of structured data generation [2/2]}
    \label{lst:system-prompt-llm-as-a-judge-02}
\end{listing}

\subsubsection{Reasoning agent prompt}\label{ax:reasoningagentprompt}

\begin{listing}[H]
    \begin{minted}[breaklines, breaksymbolleft={}]{text}
    Your task is to review a structured user input that may mention a dataset in a text. Please take your time.
    
    Carefully analyze what the text in the `mentioned_in` field explicitly means and in what context the `raw_name` is discussed. Never infer, imply, or assume, so you must exclusively rely on the text as facts. If there are multiple datasets, do the assessment individually.
    
    Plan a strategy to ensure you can maximize the chances of correctly judging and classifying whether the provided input:
    - Clearly, the `raw_name` falls under the concept of a data/dataset and not by extension or implicitly.
    - Whether the raw_name is actually in the `mentioned_in`.
    - Whether the harmonized_name (if present) is actually in the `mentioned_in`. If not found, remove it from the output.
    - The `raw_name` is `properly_named` (e.g., DHS, LSMS, etc.), `descriptive_but_unnamed` (administrative school records in Ghana for 2020) , or `vague_generic` (a survey data). Any of these are valid data mentions. To be sure, elaborate how you interpret these classes and use that for classifying.
    - The context concerning usage of the dataset is mentioned: is it `primary`, `supporting`, or `background`.
    
    Then, write down your strategy.
    
    After you write down your strategy, synthesize it to develop a rubric of what qualifies as a dataset, which you must use to base your judgment.
    
    Incorporate a devil's advocate review as part of your strategy. If the review shows inconsistency, update accordingly. Do not reason based on assumption, inference, or implicit thinking.  Relationships do not count as a dataset; for example, the producer is not a dataset.
    
    Execute the strategy, **step by step**, and write an analysis of how you interpret the `raw_name` in the context of the `mentioned_in`.
    
    If your analysis results in the `raw_name` being a dataset, set the `valid` field to `true`, otherwise, set it to `false`. In both cases, return the result of your analysis focusing on the `raw_name` in the `reason` field. If it is invalid, set the `specificity` and `context` to null.
    
    ALWAYS WRITE A DEVIL'S ADVOCATE REVIEW AFTER THE ANALYSIS BEFORE CONCLUDING.
    
    After you write your analysis, your output must repeat the input with the `specificity`, `context`, `valid` and `invalid_reason` values replaced accordingly in the same level as the corresponding `raw_name`. IMPORTANT: the final output must be between these tags <OUTPUTDATA>```json<the output must be here>```</OUTPUTDATA>
    \end{minted}
    \caption{System prompt used to characterize the reasoning agent.}
    \label{lst:system-prompt-reasoning-agent}
\end{listing}

\section{Reasoning agent example}

\begin{listing}[H]
    \begin{minted}[breaklines, breaksymbolleft={}]{text}
    {
      "mentioned_in": "In Global warming of 1.5 \nC. An IPCC\nSpecial Report on the Impacts of Global Warming of 1.5C above Pre-Industrial Levels and Related Global Greenhouse\nGas Emission Pathways, in the Context of Strengthening the Global Response to the Threat of Climate Change,\nSustainable Development, and Efforts to Eradicate Poverty; The Intergovernmental Panel on Climate Change:\nGeneva, Switzerland, 2018. 9.",
      "datasets": [
        {
          "raw_name": "IPCC Special Report on the Impacts of Global Warming of 1.5C",
          "harmonized_name": "IPCC Special Report on the Impacts of Global Warming of 1.5C",
          "acronym": "IPCC",
          "producer": "Intergovernmental Panel on Climate Change",
          "year": "2018"
        }
      ],
      "source": "b71b859da04440fe5f61613da6b223db9a74cf9c",
      "page": 11
    }

    \end{minted}
    \caption{A candidate data mention extracted from a page in a research paper.}
    \label{lst:system-prompt-reasoning-example}
\end{listing}

\begin{listing}[H]
    \begin{minted}[breaklines, breaksymbolleft={}]{text}
    ### Strategy
    
    1. **Identify the Dataset**: Determine if the `raw_name` explicitly refers to a dataset or if it is merely a title of a report or publication. A dataset should be a collection of data points or observations, while a report may summarize findings without being a dataset itself.
    
    2. **Check for Mention**: Verify if the `raw_name` is explicitly mentioned in the `mentioned_in` field. This includes checking for exact matches or clear references.
    
    3. **Harmonized Name Check**: If a `harmonized_name` is provided, ensure it is also explicitly mentioned in the `mentioned_in` field. If not, it will be excluded from the output.
    
    4. **Classify the Naming**: Classify the `raw_name` into one of the three categories:
       - **Properly Named**: Recognizable acronyms or titles (e.g., IPCC).
       - **Descriptive but Unnamed**: Descriptions of datasets without specific titles (e.g., administrative school records).
       - **Vague Generic**: General terms that do not specify a dataset (e.g., survey data).
    
    5. **Context of Usage**: Determine the context in which the dataset is used:
       - **Primary**: The main focus of the research or report.
       - **Supporting**: Provides additional information but is not the main focus.
       - **Background**: Contextual information that supports the main findings but is not directly analyzed.
    
    6. **Devil's Advocate Review**: After the initial analysis, review the findings critically to ensure there are no inconsistencies or overlooked details.

    \end{minted}
    \caption{Reasoning agent output [1/3].}
    \label{lst:system-prompt-reasoning-agent-01}
\end{listing}

\begin{listing}[H]
    \begin{minted}[breaklines, breaksymbolleft={}]{text}

    ### Rubric for Dataset Qualification
    - A dataset must be a collection of data points or observations.
    - The `raw_name` must be explicitly mentioned in the `mentioned_in` field.
    - The `harmonized_name` must also be explicitly mentioned if present.
    - Classifications must be based on clear definitions without assumptions.
    - The context of usage must be clearly identified.
    
    ### Analysis of the Input
    
    1. **Dataset Identification**: The `raw_name` "IPCC Special Report on the Impacts of Global Warming of 1.5C" appears to be the title of a report rather than a dataset. It does not represent a collection of data points but rather summarizes findings related to global warming.
    
    2. **Mention Check**: The `raw_name` is indeed mentioned in the `mentioned_in` field, but it is in the context of a report rather than a dataset.
    
    3. **Harmonized Name Check**: The `harmonized_name` is identical to the `raw_name` and is also mentioned in the `mentioned_in` field. However, since the `raw_name` is not a dataset, the harmonized name's presence does not change the classification.
    
    4. **Classification**: The `raw_name` is classified as **properly named** since it is a recognized title. However, it does not qualify as a dataset.
    
    5. **Context of Usage**: The context of the `raw_name` is primarily as a report on climate change impacts, which does not fit into the categories of primary, supporting, or background in terms of dataset usage.
    
    ### Devil's Advocate Review
    Upon reviewing the analysis, it is clear that while the `raw_name` is properly named, it does not meet the criteria for being classified as a dataset. The context of its mention reinforces that it is a report summarizing findings rather than a dataset. Therefore, the initial conclusion stands.

    \end{minted}
    \caption{[Continued] Reasoning agent output [2/3].}
    \label{lst:system-prompt-reasoning-agent-02}
\end{listing}

\begin{listing}[H]
    \begin{minted}[breaklines, breaksymbolleft={}]{text}

    ### Final Output
    ```json
    {
      "mentioned_in": "In Global warming of 1.5 \nC. An IPCC\nSpecial Report on the Impacts of Global Warming of 1.5C above Pre-Industrial Levels and Related Global Greenhouse\nGas Emission Pathways, in the Context of Strengthening the Global Response to the Threat of Climate Change,\nSustainable Development, and Efforts to Eradicate Poverty; The Intergovernmental Panel on Climate Change:\nGeneva, Switzerland, 2018. 9.",
      "datasets": [
        {
          "raw_name": "IPCC Special Report on the Impacts of Global Warming of 1.5C",
          "harmonized_name": "IPCC Special Report on the Impacts of Global Warming of 1.5C",
          "acronym": "IPCC",
          "producer": "Intergovernmental Panel on Climate Change",
          "year": "2018",
          "specificity": null,
          "context": null,
          "valid": false,
          "invalid_reason": "The raw_name is a report title and does not represent a dataset."
        }
      ],
      "source": "b71b859da04440fe5f61613da6b223db9a74cf9c",
      "page": 11
    }
    ```

    \end{minted}
    \caption{[Continued] Reasoning agent output [3/3].}
    \label{lst:system-prompt-reasoning-agent-03}
\end{listing}

\end{document}